\documentclass[sigconf]{acmart}
\usepackage{graphicx}
\usepackage{array}
\usepackage{float}
\usepackage{booktabs} %
\usepackage{tabularx} %
\usepackage{geometry}
\usepackage{colortbl} 

\usepackage{multirow}    %
\usepackage{caption}     %
\usepackage{siunitx}
\newcommand{\Softmax}{\mathop{\rm{Softmax}}}
\AtBeginDocument{%
  }

\copyrightyear{2024}
\acmYear{2024}
\setcopyright{rightsretained}
\acmConference[HuMob'24]{2nd ACM SIGSPATIAL International Workshop on the Human Mobility Prediction Challenge}{October 29-November 1, 2024}{Atlanta, GA, USA}
\acmBooktitle{2nd ACM SIGSPATIAL International Workshop on the Human Mobility Prediction Challenge (HuMob'24), October 29-November 1, 2024, Atlanta, GA, USA}
\acmDOI{10.1145/3681771.3699910}
\acmISBN{979-8-4007-1150-3/24/10}

\newcommand{\sys}{{\rm ST-MoE-BERT}}

\newcommand{\syb}{{\rm \textbf{ST-MoE-BERT}}}
\begin{document}

\title{ST-MoE-BERT: A Spatial-Temporal Mixture-of-Experts Framework for Long-Term Cross-City Mobility Prediction
}

\author{Haoyu He}
\affiliation{%
  \institution{Northeastern University}
  \city{Boston}
  \state{MA}
  \country{USA}}
\email{he.haoyu1@northeastern.edu}

\author{Haozheng Luo}
\affiliation{%
  \institution{Northwestern University}
  \city{Evanston}
  \state{IL}
  \country{USA}}
\email{robinluo2022@u.northwestern.edu}

\author{Qi R. Wang}
\affiliation{%
 \institution{Northeastern University}
 \city{Boston}
 \state{MA}
 \country{USA}}
\email{q.wang@northeastern.edu}

\renewcommand{\shortauthors}{He et al.}

\begin{abstract}
  Predicting human mobility across multiple cities presents significant challenges due to the complex and diverse spatial-temporal dynamics inherent in different urban environments.
In this study, we propose a robust approach to predict human mobility patterns called \syb. 
Compared to existing methods, our approach frames the prediction task as a spatial-temporal classification problem.
Our methodology integrates the Mixture-of-Experts architecture with BERT model to capture complex mobility dynamics and perform the downstream human mobility prediction task.
Additionally, transfer learning is integrated to solve the challenge of data scarcity in cross-city prediction.
We demonstrate the effectiveness of the proposed model on GEO-BLEU and DTW, comparing it to several state-of-the-art methods.
Notably, \sys achieves an average improvement of 8.29\%.
\footnote{Code is available at \href{https://github.com/he-h/HuMob}{\textbf{Github}}.}

\end{abstract}



\maketitle

\section{Introduction}
\label{intro}
Human mobility is a cornerstone of societal functioning, underpinning economic activities, urban development, and social interactions \cite{wang2018urban}. Understanding human travel patterns is essential for optimizing transportation systems, enhancing urban planning, and managing public health initiatives~\cite{he2022percolation, barbosa2018human}. In recent years, human mobility prediction has emerged as a critical component in the development of intelligent urban systems \cite{luca2021survey}. Accurate forecasting of individual and population movement patterns enables proactive decision-making, improves resource allocation, and enhances the responsiveness of services to dynamic urban environments.

However, predicting human mobility poses several significant challenges. Firstly, mobility data are often sparse and unevenly distributed spatially and temporarily, making it difficult to capture comprehensive movement patterns \cite{yabe2024enhancing}. Secondly, human behaviors are inherently complex and influenced by a myriad of factors such as social interactions \cite{yan2017universal} and environmental changes \cite{bontorin2024mixing}, which complicates the modeling of their dependencies \cite{wang2011human}. Lastly, transferring predictive models between different cities is challenging due to the heterogeneity in mobility patterns, urban infrastructure, and demographic characteristics \cite{dalziel2013human}.

\begin{figure*}[htp]
    \centering
    \includegraphics[width=\linewidth]{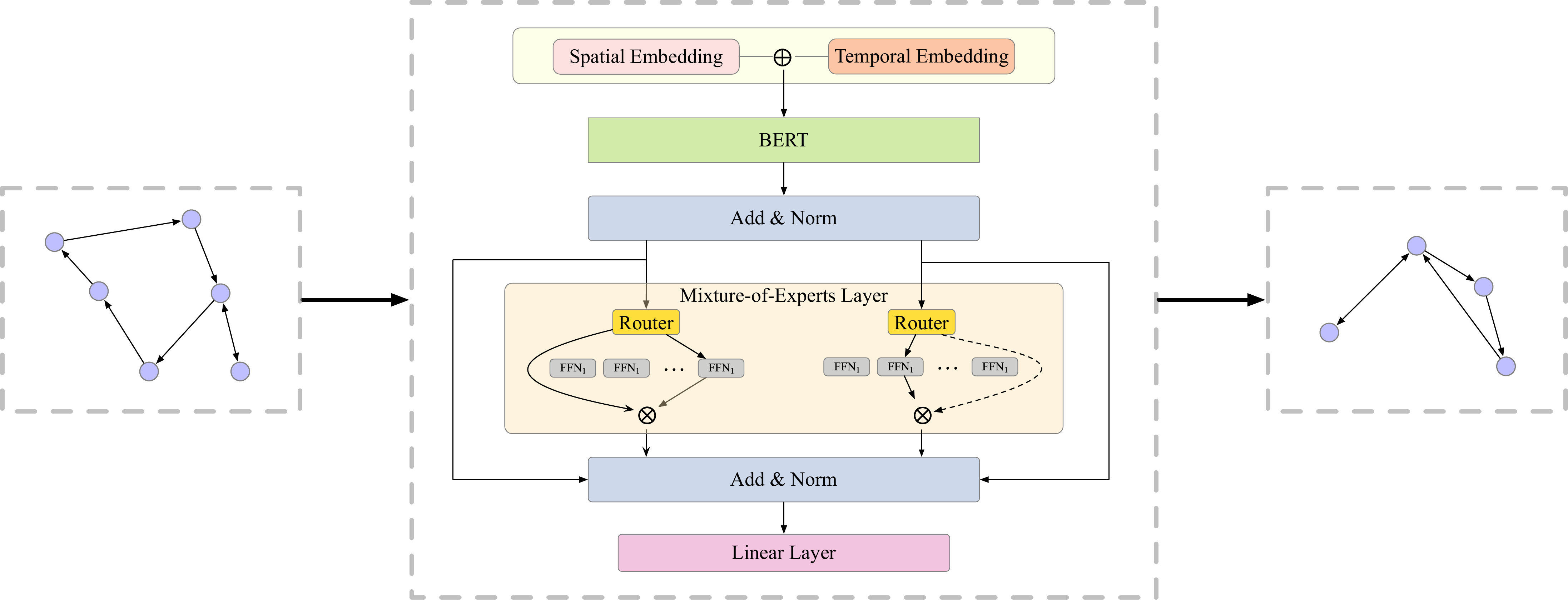}
     \vspace{-0.1in}
    \caption{Overview of ST-MoE-BERT workflow. \textnormal{Input trajectories are first encoded with spatial and temporal embeddings, then combined and fed into the BERT model. The resulting representations are passed through a Mixture-of-Experts (MoE) layer consisting of eight feedforward network (FFN) experts. The MoE router selects the top two experts for each input, and their outputs are merged in a final linear layer to generate the predicted locations.}}
    \label{fig:pipeline}
\end{figure*}

To address these challenges, we propose \syb (\underline{S}patial-\underline{T}emporal \underline{M}ixture-\underline{o}f-\underline{E}xperts with \underline{BERT}), an innovative framework that integrates transformer-based architectures with a Mixture-of-Experts (MoE) layer \cite{jacobs1991adaptive}. The structure of our proposed framework is shown in \autoref{fig:pipeline}. \sys leverages the sequence modeling capabilities of BERT \cite{devlin2018bert} and the specialized expertise of MoE networks to capture both general and city-specific mobility patterns. Furthermore, our transfer learning strategy enables the model to adapt knowledge from data-rich cities to those with limited datasets, thereby enhancing prediction accuracy across multiple urban environments. Our key contributions are as follows.

\begin{itemize}
    \item We introduce \sys, a unique transformer-based method that combines BERT with an MoE layer to effectively address the long-term cross-city mobility prediction problem.
    \item We develop a transfer learning strategy that employs differential learning rates, thereby enhancing prediction accuracy in data-scarce environments and facilitating adaptation to diverse spatial distributions.
    \item Experimentally, we demonstrate that \sys outperforms the state-of-the-art methods on the long-term cross-city prediction task with an average improvement of 8.29\%.
\end{itemize}

\section{ST-MoE-BERT}
\label{method}
In this section, we present the proposed ST-MoE-BERT framework for long-term human mobility prediction.

\paragraph{\textbf{Problem Setup.}}
Given a sequence of historical mobility records $\mathcal{X} = \{x_1, x_2, \ldots, x_T\}$, where each $x_t \in \mathcal{L}$ represents the user's location at time step $t$, the goal is to predict the future mobility trajectory $\mathcal{Y} = \{y_{T+1}, y_{T+2}, \ldots, y_{T+H}\}$, where each $y_t \in \mathcal{L}$ denotes the user's location at time step $t$. 
Here, $T$ is the length of the historical sequence, and $H$ is the prediction horizon.

To achieve this, we aim to train a predictor $f: \mathcal{L}^T \rightarrow \mathcal{L}^H$ that minimizes the cross-entropy loss between the predicted and ground-truth trajectories on the training set of $N$ trajectories. 
The loss function $\mathcal{J}$ is defined as:
\begin{align*}
    \mathcal{J} = -\sum_{t=T+1}^{T+H} \sum_{i=1}^{N} \mathbb{I}(y_{t,i}) \log \hat{y}_{t,i},
\end{align*}
where $\mathbb{I}(y_{t,i})$ equals $1$ if the prediction for sample $i$ at time $t$ is correct, otherwise $0$; and $\hat{y}_{t,i}$ represents the predicted probability distribution over all possible locations for sample $i$ at time $t$.

In this classification set-up, each location is treated as a distinct class, and the predictor $f$ outputs a probability distribution over the classes for each future time step.

\paragraph{\textbf{Motivating Example.}}
Recently, \citet{liu2023itransformer} propose modeling multivariate correlations by treating independent time series as tokens using self-attention mechanism. 
We use this observation as a starting point by considering our spatial-temporal foundation model with a combination embedding of weekday, day, time of day, and weekend.
Since our problem is a long-term classification problem, we use the autoregressive model BERT \cite{devlin2018bert} to model the temporal dependencies in the mobility data.
To capture diverse and long-term mobility patterns, we incorporate a Mixture of Experts (MoE) layer \cite{jacobs1991adaptive} into our model architecture. 
The MoE layer consists of multiple specialized expert networks, each designed to model distinct aspects of human mobility. 
A gating network dynamically assigns the input data to these experts, allowing the model to adaptively focus on the relevant patterns for each prediction task.

\begin{figure*}[htp]
    \centering
    \includegraphics[width=.9\linewidth]{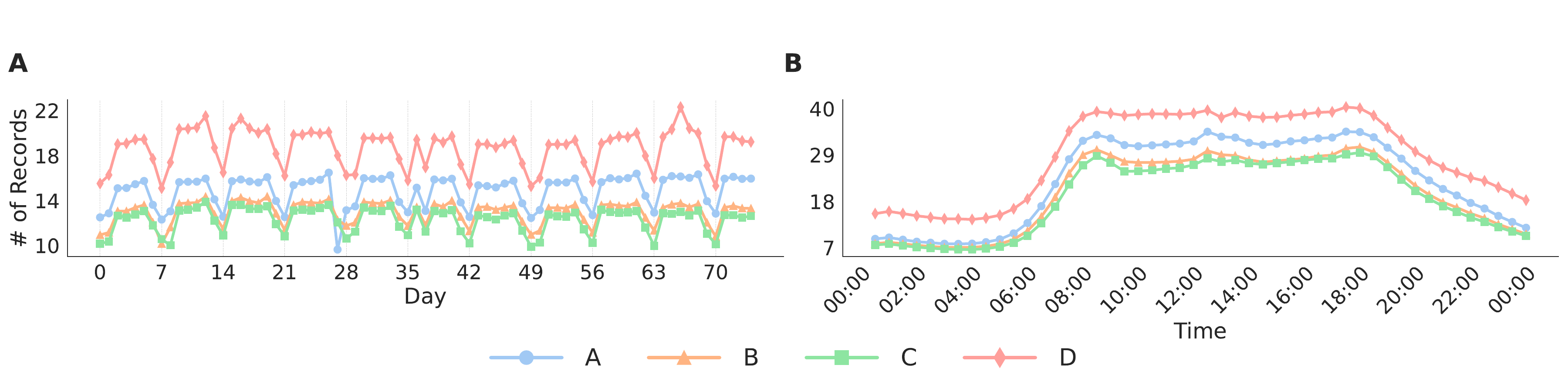}
     \vspace{-0.1in}
    \caption{Average Mobility Records Per User Over Different Time Scales. A: \textnormal{Average number of records per user over a 75-day period in four cities, displaying a weekly cyclical pattern with higher records from Monday to Friday. } B: \textnormal{Average half-hourly records per user for a single day, showing peak activity between 8 AM and 6 PM.}}
    \label{fig:trend}
\end{figure*}

\subsection{Model Structure}
In this section, we introduce our proposed \sys framework. 
The primary architecture combines BERT with a Mixture of Experts (MoE) layer.

\paragraph{\textbf{BERT.}}
To effectively capture the intricate temporal dependencies in human mobility data, we integrate the transformer architecture \cite{vaswani2017attention} into our model. 
Specifically, we employ BERT~\cite{devlin2018bert} as the backbone, leveraging its robust self-attention mechanisms to understand the relationship between a user's historical and future trajectories. Additionally, BERT excels at encoding contextual information from input sequences, and thus the model is able to discern complex movement patterns over time.

The self-attention mechanism, defined as:
\begin{align*} 
    \text{Attention}(X) = \Softmax \left(\frac{QK^T}{\sqrt{d_k}}\right)V,
\end{align*}
computes attention weights by projecting the input sequence into query ($Q$), key ($K$), and value ($V$) matrices. This process allows the model to assign varying levels of importance to different time steps in historical sequences to focus on the most relevant locations when predicting future movements. The $\Softmax$ function ensures that the attention weights are normalized to effectively aggregate information across the sequence.

Furthermore, BERT utilizes a \textit{[CLS]} token to capture global information from the entire input sequence. We leverage this \textit{[CLS]} token as the input for the subsequent layer, ensuring that both local and global temporal contexts are effectively utilized.

\paragraph{\textbf{Mixture of Experts.}}
To effectively capture diverse and long-term mobility patterns across multiple cities, we incorporate a Mixture of Experts (MoE) layer \cite{jacobs1991adaptive} into our model architecture. The MoE layer comprises multiple specialized expert networks, each designed to model distinct aspects of mobility patterns, varying spatial regions, or temporal behaviors. A gating network dynamically assigns the input data to these experts, allowing the model to adaptively focus on the relevant patterns for each prediction task.

Mathematically, for an input feature vector $\mathbf{x} \in \mathbb{R}^d$, the MoE layer computes the output as:
\begin{align*}
    \text{MoE}(\mathbf{x}) = \sum_{k=1}^K g_k(\mathbf{x}) f_k(\mathbf{x}),
\end{align*}
where $f_k(\mathbf{x})$ represents the output of the $k$-th expert, and $g_k(\mathbf{x})$ is the gating network's probability for expert $k$, satisfying $\sum_{k=1}^K g_k(\mathbf{x}) = 1$ and $g_k(\mathbf{x}) \geq 0$. The gating probabilities $g_k(\mathbf{x})$ are derived using a softmax function:
\begin{align*}
    g_k(\mathbf{x}) = \frac{\exp(\mathbf{w}_k^\top \mathbf{x} + b_k)}{\sum_{j=1}^K \exp(\mathbf{w}_j^\top \mathbf{x} + b_j)}.
\end{align*}
In the context of human mobility prediction, the MoE layer allows the model to specialize different experts to distinct mobility patterns, such as varying user behaviors across different cities or temporal trends. This specialization enhances the model's ability to generalize across diverse mobility scenarios. Furthermore, the gating mechanism ensures that the model dynamically selects the most relevant experts based on the input data, thereby adapting to evolving mobility trends and improving prediction accuracy in multi-city settings.

\subsection{Transfer Learning}
Transfer learning \cite{weiss2016survey} involves taking a pre-trained model on a large dataset and fine-tuning it for a different but related task. \citet{furst2022cloob} offers a theoretical analysis of how transfer learning operates within transformer architectures, highlighting the benefits of leveraging pre-trained models on related tasks. In human mobility prediction, transfer learning enables the adaptation of the model's learned representations to new urban environments, thereby leveraging knowledge from previously studied cities to enhance prediction accuracy in others.

Drawing inspiration from recent advancements in dense associative memory models \cite{furst2022cloob, hu2024outlier}, where the attention mechanism in transformers is conceptualized as a form of associative memory, we propose a transfer learning strategy tailored for multi-city mobility prediction. Specifically, we pretrain our model on a large-scale mobility dataset from one city and subsequently fine-tune it on smaller datasets from other cities. This approach allows the model to capture both general and city-specific mobility patterns.

To effectively learn the distinct spatial information of each city, we employ different learning rates during fine-tuning. Specifically, we set the learning rate for the location embeddings to be ten times higher than the base learning rate used for the other model parameters. This higher learning rate enables the location embeddings to rapidly adapt and capture the unique spatial characteristics of the target city. Conversely, we apply a smaller learning rate to the rest of the model parameters to preserve the general knowledge acquired during pretraining. This strategy facilitates the transfer of knowledge from one city to another, improves prediction accuracy, and addresses uneven data distributions across cities.

\section{Experiments}
\label{experiments}

\begin{table*}[htbp]
    \centering
    \caption{Comparison of ST-MoE-BERT with Benchmark Methods for Cross-city Prediction. \textnormal{We perform experiments on cross-city prediction tasks using two baseline models. Evaluation metrics include GEO-BLEU, DTW, and accuracy across four cities. 'ST-MoE-BERT w/o PT' refers to the model trained directly on each city without pretraining on city A. The best results are shown in bold. In the majority of configurations, ST-MoE-BERT consistently outperforms all baselines.}}
    \vspace{-0.1in}
    \resizebox{\textwidth}{!}{
    \begin{tabular}{lcccccccccccc}
    \toprule
    Method  &  \multicolumn{3}{c}{A} & \multicolumn{3}{c}{B} & \multicolumn{3}{c}{C} & \multicolumn{3}{c}{D}\\
    \cmidrule(lr){2-4} \cmidrule(lr){5-7} \cmidrule(lr){8-10} \cmidrule(lr){11-13}
     &  GEO-BLEU $\uparrow$ & DTW $\downarrow$ & Acc. $\uparrow$ & GEO-BLEU $\uparrow$ & DTW  $\downarrow$ & Acc. $\uparrow$ & GEO-BLEU $\uparrow$ & DTW  $\downarrow$ & Acc. $\uparrow$ & GEO-BLEU $\uparrow$ & DTW  $\downarrow$ & Acc. $\uparrow$\\
    \midrule
       \cellcolor{gray!15} HF & \cellcolor{gray!15} 0.266 & \cellcolor{gray!15} 80.3 & \cellcolor{gray!15} 20.4\% & \cellcolor{gray!15} 0.265 & \cellcolor{gray!15} 56.4 & \cellcolor{gray!15} 21.0\% & \cellcolor{gray!15} 0.251 & \cellcolor{gray!15} 42.4 & \cellcolor{gray!15} 20.8\% & \cellcolor{gray!15} 0.295 & \cellcolor{gray!15} 80.0 & \cellcolor{gray!15} 21.0\% \\
        BERT & 0.256 & 35.7 & 23.8\% & 0.284 & \textbf{20.6} & 27.0\% & 0.253 & 65.6 & 18.2\% & 0.253 & 65.6 & 18.2\%\\
    \midrule
        \cellcolor{gray!15} ST-MoE-BERT w/o PT & \cellcolor{gray!15} \textbf{0.286} &\cellcolor{gray!15}  \textbf{30.2} & \cellcolor{gray!15} \textbf{27.9\%} &\cellcolor{gray!15}  0.286 & \cellcolor{gray!15} 28.2 &\cellcolor{gray!15} 27.5\% & \cellcolor{gray!15} 0.294 & \cellcolor{gray!15} 20.7 & \cellcolor{gray!15} 27.9\% & \cellcolor{gray!15} 0.250 &\cellcolor{gray!15}  67.6 & \cellcolor{gray!15} 21.4\% \\
        ST-MoE-BERT & - & - & - & \textbf{0.297}  & 29.3 & \textbf{28.7\%} & \textbf{0.297}  & \textbf{19.7} & \textbf{28.9\%} & \textbf{0.300}  & \textbf{48.1} & \textbf{26.5\%}\\
    \bottomrule
    \end{tabular}
    }
    \label{tab:results}
\end{table*}

In this section, we conduct a series of experiments to demonstrate the performance and transferability of \sys.

\paragraph{\textbf{Models.}}
In our experiments, we validate our approach using \sys. We adopt the BERT model \footnote{\url{https://huggingface.co/google-bert/bert-base-uncased}} with 110 million parameters. The input sequence length is set to 240, and the prediction horizon is 48. We pretrain the model using the masked language modeling (MLM) technique on mobility data from city A and then fine-tune it on mobility data from cities B, C, and D.
The hyperparameters of \sys are detailed in \autoref{tab:hyperparameters}.

\paragraph{\textbf{Datasets.}}
We evaluate our method on human mobility data from four metropolitan areas, labeled as cities A, B, C, and D, as provided by \citet{yabe2024yjmob100k}. Each city is divided into a 200 $\times$ 200 grid, where each cell corresponds to a 500-meter by 500-meter area. The mobility datasets cover a 75-day period, with individual movements recorded at 30-minute intervals within these grid cells.

\paragraph{\textbf{Evaluation Metrics.}}
To evaluate the performance of predicting future mobility trajectories, we employ three metrics: Accuracy, GEO-BLEU~\cite{shimizu2022geo}, and Dynamic Time Warping (DTW)~\cite{muller2007dynamic}. Accuracy quantifies the percentage of correctly predicted grid cells in future trajectories. GEO-BLEU is a metric that accounts for both precision and temporal alignment between the predicted and actual trajectories. DTW measures the alignment and distance between the predicted and true trajectories and is commonly used in long-term mobility prediction tasks \cite{he2023forecasting}. Further details on GEO-BLEU and DTW are provided in \autoref{appendix_metric}.

\paragraph{\textbf{Data Processing.}}
We preprocess the human mobility data by categorizing the grid cells into 40,000 distinct classes. The dataset is then divided into training and testing sets, with the first 60 days allocated for training and the following 15 days for testing. To assess the percentage of missing data, we analyze the completeness of the data in each city, as shown in \autoref{fig:missing}. In our experiments, we focus our analysis exclusively on data where each time window has corresponding location records. Additionally, we examine the daily and hourly trends of data records across cities, as presented in \autoref{fig:trend}.

\subsection{Cross-City Prediction with ST-MoE-BERT}
To evaluate the efficiency of our method on cross-city prediction, we compare \sys with baseline models on predictions for cities B, C, and D. Each evaluation is conducted three times using different random seeds, and we report the average performance for each metric.

\paragraph{\textbf{Baselines.}}
To evaluate the performance of our method, we use Historical Frequency (HF), naive BERT, and \sys without pretraining to assess the efficiency of \sys on cross-city prediction. HF predicts future locations using historical visit patterns based on time and weekday.

\paragraph{\textbf{Results.}}
As shown in \autoref{tab:results}, we observe an average improvement of 10.30\% in GEO-BLEU, 15.50\% in DTW, and 21.40\% in accuracy across three cities. It is evident that \sys outperforms all baseline models in predicting mobility for cities B, C, and D. HF scores high on GEO-BLEU by matching frequent past locations but may predict positions far from the true grid. 
In most configurations, \sys without pretraining (w/o PT) outperforms the vanilla BERT model, highlighting the effectiveness of the MoE mechanism in capturing complex relationships across different cities. However, there are exceptions where the naive BERT model outperforms \sys without pretraining, such as the GEO-BLEU score in city D. One possible reason is that the naive BERT model is more attuned to the distinct data distribution of city D, which differs from the other cities.

\subsection{Ablation Study: Impact of Transfer Learning on Prediction Performance}
We conduct experiments to assess the impact of transfer learning on the final prediction results in cross-city prediction tasks. As shown in \autoref{tab:results}, we observe an average improvement of 8.29\% in GEO-BLEU, 9.90\% in DTW, and 10.76\% in accuracy across the three cities. It is evident that \sys, when utilizing knowledge from other cities, enhances the robustness of cross-city predictions. 
The only exception, where \sys without transfer learning outperforms the model with transfer learning, is the DTW score in city B.

\section{Discussion}
\label{discussion}
We introduce \sys for predicting long-term human mobility patterns across cities. \sys utilizes a transformer-based architecture with an MoE layer to capture complex spatio-temporal dependencies in mobility data. To enhance model robustness in target cities with limited data, we apply transfer learning by pretraining the model on city A. Our experiments show that \sys improves prediction accuracy by an average of 8.29\% compared to state-of-the-art models. These results highlight the significance of combining transformer architectures with MoE layers and a strong transfer learning strategy for better long-term cross-city human mobility prediction.

\textbf{Limitation and Future Work.} Despite its strengths, \sys relies on a large-scale dataset for pre-training which may limit the model’s applicability in scenarios where such comprehensive data is unavailable. Future work can address these limitations by exploring the integration of LLMs, which have shown promise in capturing intricate and multifaceted aspects of human behavior.

\begin{acks}
He and Wang acknowledge the support from the U.S. National Science Foundation (NSF) under Grant No. 2125326, 2114197, 2228533, and 2402438. The authors are grateful for the support of NSF. Any opinions, findings, conclusions, or recommendations expressed in the paper are those of the author and do not necessarily reflect the views of the funding agencies.
\end{acks}

\clearpage
\bibliographystyle{ACM-Reference-Format}
\bibliography{reference}
\definecolor{darkred}{rgb}{0.6, 0, 0}

\hypersetup{
    colorlinks=true,
    linkcolor=darkred,  %
    urlcolor=darkred    %
}
\appendix
\label{appendix}

\section{Related Work}
\label{related_work}
In this section, we present a concise overview of human mobility prediction and the application of foundation models for time series classification.

\textbf{Human Mobility Prediction.}
Researchers have applied physics-based models to understand human mobility, depicting movements as scale-free Lévy flights with distances following a truncated power-law distribution \cite{brockmann2006scaling, rhee2011levy}. Patterns such as frequently revisited locations align with Zipf's Law \cite{gonzalez2008understanding}, while the number of unique places visited grows sublinearly \cite{song2010limits}.
Although Markov models have been used to predict future locations based solely on the current state \cite{qiao2018hybrid}, they struggle to capture the complex spatio-temporal dependencies and varied travel behaviors in human mobility.

With the advancement of deep learning, more sophisticated models have been developed to address these limitations. Recurrent neural networks, such as STRNN \cite{liu2016predicting}, were applied to mobility prediction, effectively capturing temporal dependencies in sequential mobility data. Building upon this, the attention mechanism was introduced in DeepMove \cite{feng2018deepmove}, allowing the model to focus on the most relevant part of a user's trajectories. MobTCast \cite{xue2021mobtcast} employed context-aware transformer to effectively model users' social and geographical interactions. More recently, cross-city prediction approaches \cite{wang2024cola} have highlighted the transferability of mobility knowledge across different urban areas by leveraging shared spatio-temporal patterns. A novel direction involves the exploration of large language models (LLMs) for mobility prediction \cite{liang2024exploring, beneduce2024large}. These studies utilize the vast capabilities of LLMs, along with prompt-based learning, to effectively capture latent mobility patterns and provide more accurate, context-specific predictions across diverse environments.

\textbf{Foundation Models for Time Series Classification.}
The advent of Transformer-based architectures, initially introduced for language translation tasks \cite{ devlin2018bert,vaswani2017attention}, has paved the way for the development of foundation models across diverse domains. These architectures are capable of capturing complex relationships within various types of data, thereby leading to significant advancements in fields, such as safety \cite{luo2024decoupled,yu2024enhancing} and prompting \cite{lester2021power, gao2020making}.
Meanwhile, they have demonstrated exceptional performances across multiple scientific disciplines, including genomics \cite{nguyen2024hyenadna}, finance \cite{wang2023fingpt,wu2023bloomberggpt} and many others.
In the context of time series forecasting, foundation models have also been explored for time series forecasting, with approaches built on existing LLMs \cite{jin2023time, garza2023timegpt} showing promising results on tasks like forecasting, imputation, and anomaly detection. 
However, current time-series foundation models \cite{jin2023time,das2023decoder,gruver2024large} rely on patch-based embedding mechanisms, which are inadequate for capturing the intricate spatio-temporal dependencies in human mobility data. 
In contrast to TimeLLM \citep{jin2023time}, \sys is specifically designed to model these complex spatio-temporal relationships, offering more accurate and context-aware predictions for human mobility.

\section{Experiment Setting}
\subsection{Hyperparameters of ST-MoE-BERT}
The hyperparameters of \sys are presented in \autoref{tab:hyperparameters}.

\begin{table}[htp]
\centering
\caption{Hyperparameters of the Pretrained and Fine-Tuned Models}
\label{tab:hyperparameters}
\resizebox{0.5\textwidth}{!}{
\begin{tabular}{@{} >{\raggedright\arraybackslash}p{6cm} >{\centering\arraybackslash}p{3cm} @{}}
\toprule
\multicolumn{2}{c}{\textbf{Pretrained Model}} \\
\midrule
Learning Rate                         & 0.0003           \\
Weight Decay                            & 0.001            \\
\midrule
Number of Hidden Layers               & 12             \\
Hidden Size                           & 768            \\
Number of Attention Heads             & 16             \\
Number of Experts                     & 8              \\
Dropout                    & 0.1              \\
\midrule
Day Embedding Size                    & 64             \\
Time Embedding Size                   & 64             \\
Day of Week Embedding Size            & 64             \\
Weekday Embedding Size                & 32             \\
Location Embedding Size               & 256            \\
\midrule
\multicolumn{2}{c}{\textbf{Fine-Tuned Model}} \\
\midrule
Learning Rate                         & 0.00005           \\
Location Embedding Learning Rate      & 0.0005           \\
\bottomrule
\end{tabular}
}
\end{table}

\subsection{Data Completeness Rate}

\begin{figure}[htbp]
    \centering
    \includegraphics[width=0.5\linewidth]{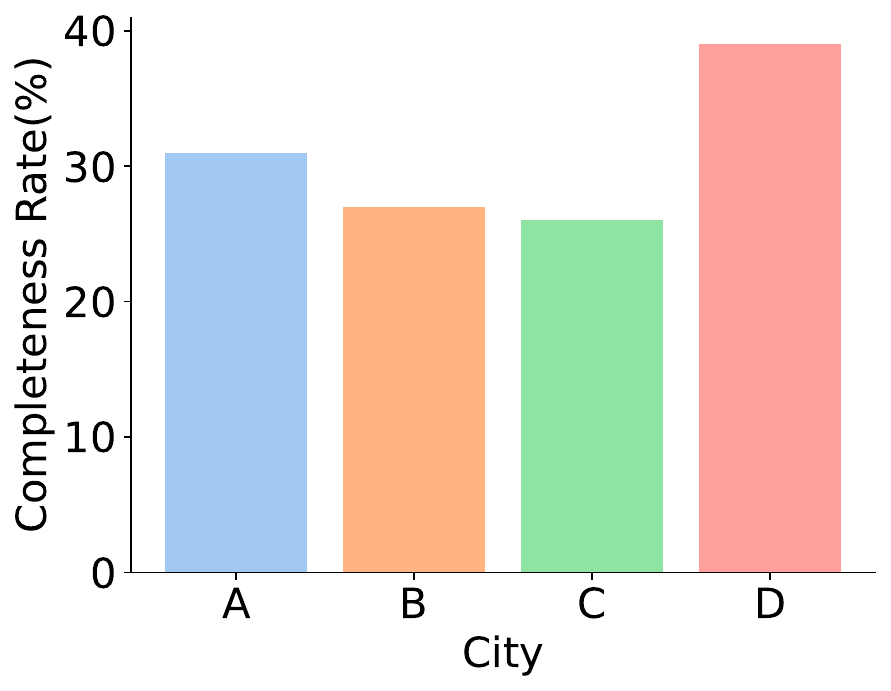}
    \caption{\textbf{The completeness rate for each city.}}
    \label{fig:missing}
\end{figure}

\subsection{Distribution of Individuals Across Cities}

\begin{table}[H]
\centering
\caption{Number of users with movement data across cities A, B, C, and D.}
\begin{tabular}{lcccc}
\toprule
\textbf{City} & \textbf{A} & \textbf{B} & \textbf{C} & \textbf{D} \\
\midrule
\textbf{Number of Users} & 100,000 & 25,000 & 20,000 & 6,000 \\
\bottomrule
\end{tabular}
\label{tab:number_of_users}
\end{table}

\subsection{Evaluation metrics}
\label{appendix_metric}

Given two sequences, \(X = (x_1, x_2, \ldots, x_n)\) and \(Y = (y_1, y_2, \ldots, y_m)\):

\textbf{GEO-BLEU:} GEO-BLEU extends the BLEU metric to geographic data. It evaluates n-gram similarities between sequences using the formula:
\[
\text{GEO-BLEU} = BP * \exp \left( \sum_{n=1}^N w_n \log q_n \right)
\]
where \(BP\) is the brevity penalty, \(w_n\) are the weights for each n-gram level, and \(q_n\) is the geometric mean of the n-gram similarities between the sequences.

\textbf{DTW:} DTW measures the similarity between \(X\) and \(Y\) by finding the optimal alignment that minimizes the sum of Euclidean distances between corresponding elements. It is defined as:
\[
\text{DTW}(X, Y) = \min_{c} \left( \sum_{(i,j) \in c} \text{dist}(x_i, y_j) \right)
\]
where \(c\) denotes the alignment path and \(\text{dist}(x_i, y_j)\) is the Euclidean distance between \(x_i\) and \(y_j\).

\end{document}